\pdfoutput=1
\documentclass[11pt, a4paper]{article}
\usepackage[margin=0.96in]{geometry}

\usepackage[]{emnlp2021}

\usepackage{times}
\usepackage{latexsym}

\usepackage[T1]{fontenc}
\usepackage[utf8]{inputenc}

\usepackage{microtype}

\usepackage{amsfonts}
\usepackage{amsmath}
\usepackage{graphicx}
\usepackage{multirow}
\usepackage{pifont}
\newcommand{\xmark}{\ding{55}}%

\begin{document}
\title{Recall and Learn: A Memory-augmented Solver for Math Word Problems}

\author{Shifeng Huang \footnotemark[1] \\
 CVTE Research \\
 \texttt{huangshifeng@cvte.com} \\ \And
 Jiawei Wang \footnotemark[1] \footnotemark[2] \\ 
 CVTE Research \\
 \texttt{wangjiawei0531@gmail.com} \\ \AND
 Jiao Xu \\
 CVTE Research \\
 \texttt{xujiao@cvte.com} \\ \And
 Da Cao \\ 
 Hunan University \\ 
 \texttt{caoda0721@gmail.com} \\ \And
 Ming Yang \\
 CVTE Research \\
 \texttt{yangming@cvte.com}
 }
  
\maketitle

\renewcommand{\thefootnote}{\fnsymbol{footnote}} %将脚注符号设置为fnsymbol类型，即特殊符号表示
\footnotetext[1]{Both authors contributed equally to this research.} %对应脚注[1]
\footnotetext[2]{Corresponding author.} %对应脚注[2]

\renewcommand{\thefootnote}{\arabic{footnote}}
\setcounter{footnote}{0}

\begin{abstract}
In this article, we tackle the math word problem, namely, automatically answering a mathematical problem according to its textual description. Although recent methods have demonstrated their promising results, most of these methods are based on template-based generation scheme which results in limited generalization capability. To this end, we propose a novel human-like analogical learning method in a recall and learn manner. Our proposed framework is composed of modules of memory, representation, analogy, and reasoning, which are designed to make a new exercise by referring to the exercises learned in the past. Specifically, given a math word problem, the model first retrieves similar questions by a memory module and then encodes the unsolved problem and each retrieved question using a representation module. Moreover, to solve the problem in a way of analogy, an analogy module and a reasoning module with a copy mechanism are proposed to model the interrelationship between the problem and each retrieved question. Extensive experiments on two well-known datasets show the superiority of our proposed algorithm as compared to other state-of-the-art competitors from both overall performance comparison and micro-scope studies.
\end{abstract}

\section{Introduction}
The task of Math Word Problem (MWP) aims at automatically solving a mathematical question according to its textual description. Given a problem description, a model needs to understand the relevant quantities and reason the corresponding expression, which is a difficult task because it requires the model to learn mathematics knowledge from the labeled problem and generalize the knowledge to the unseen problems.

In fact, great efforts have been made to address the MWPs in the research community. Boosted by the proliferation of deep learning techniques, Seq2Seq-based models have been developed to solve MWPs. \citet{wang2017deep} presented a large-scale MWP dataset Math23K and proposed an RNN-based framework with a number mapping technique, which aims to generate a math template first, and then fill the extracted number from the problem into the slots of the generated template to obtain an expression. This two-stage method is widely used as a baseline by the latest papers, such as Math-EN \citep{wang2018translating}, GTS \citep{xie2019goal}, Graph2Tree \citep{zhang2020graph}, Ape \citep{zhao2020ape210k} and so on \citep{wang2019template, li2019modeling}.

\begin{figure}[t]
	\centering
	\includegraphics[width=0.48\textwidth]{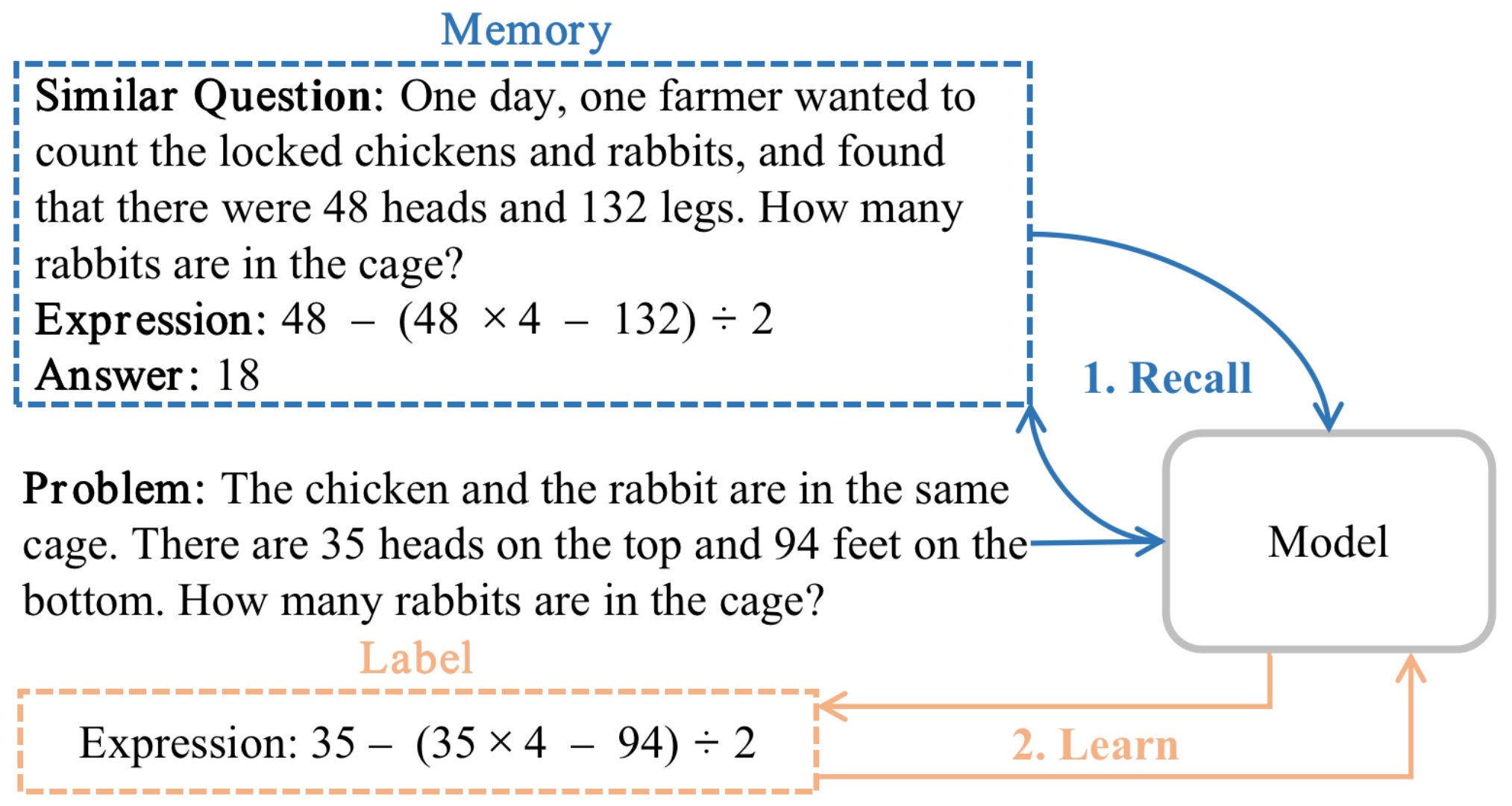}
	\vspace{-0.4cm}
	\caption{Illustration of our proposed framework for solving math word problems in a recall and learn manner.}
	\label{fig: MWP}
\end{figure}

Despite its value and significance, the math word problem has not been well addressed due to the following challenges: 1) Although promising results have been reported, the aforementioned models all use the template-based framework to solve MWPs, such a two-stage process may introduce systematic cumulative errors. In light of this, how to solve MWPs properly without using the template is a non-trivial task. 2) Furthermore, instead of learning through a single training example, the way human learn often rely on the so-called analogical learning method, which is able to explore the inherent laws between various cases and generalize them to new examples \cite{schwartz2016abcs, hope2017accelerating}. Therefore, how to combine the analogical learning method in a unified framework is worth exploring.

To address the aforementioned issues, as revealed in Figure \ref{fig: MWP}, we design a novel memory-augmented model named REAL (short for ``REcall And Learn'') to solve the MWP task in an end-to-end manner. REAL is able to recall some familiar questions that have been solved when solving a new problem, and learns to generate a similar solution in an analogical way. Specifically, REAL model first initializes a memory module by a dataset formed with questions and their expressions. When solving a problem, the memory module is utilized to retrieve the most similar questions as references according to the unsolved problem. Next, a representation module is proposed to extract item memories of the unsolved problem and the retrieved question. Thereafter, we employ an analogy module to construct relational memory based on the item memories. Finally, a reasoning module is applied to generate the expression of the unsolved problem by combining the generation and copy mechanisms. Extensive experiments show that we have achieved competitive performance on MWP task. Moreover, our proposed model is able to improve the performance by retrieving more questions, which shows the model has the ability to learn by analogy.

The main contributions of this work are summarized as follows:
\begin{itemize}
	\item{To the best of our knowledge, this is the first model that learns to solve math word problems using human-like analogical learning way.}
	\item{We develop a novel memory-augmented framework combined with the copy mechanism, REAL, to solve MWPs in a recall and learn manner, in which the model is composed of modules of memory, representation, analogy and reasoning.}
	\item{Extensive experiments are conducted on two well-known datasets, and the results showed that the REAL model not only achieves competitive performance on MWP task, but also demonstrates the unique ability of learning by analogy. Meanwhile, we have released the code to facilitate the research community.\footnote{\url{https://github.com/sfeng-m/REAL4MWP}}}
\end{itemize}

\section{Related Work}
In this section, we briefly review some literatures that are tightly related to our work, namely, math word problems and memory-augmented generative methods.
\subsection{Math Word Problems}
In the MWP task, the algorithms are designed to calculate a mathematical expression based on the textual description of mathematical problems. Therefore, the methods of natural language processing can be widely used in MWP task. Most of existing models adopt an encoder-decoder framework, where the encoder is designed as a bidirectional RNN and the decoder is designed as a unidirectional RNN. For example, \citet{wang2017deep} constructed a large dataset and proposed a Seq2Seq model that shows the superiority over previous works. \citet{wang2018translating} proposed an equation normalization technique to solve the order-duplicated problem and bracket-duplicated problem. \citet{wang2019template} designed a tree-structure model to predict the suffix expression of MWPs, which reduces the target space of the problem. \citet{xie2019goal} proposed a tree-structured gated recurrent unit as decoder, which passes the information through the expression tree in both top-down and bottom-up manners. \citet{zhang2020graph} proposed a graph encoder to enrich the quantity representations in the problem, and decode the expression by a tree structure decoder. \citet{zhao2020ape210k} presented a new large-scale and template-rich MWP dataset Ape210K and proposed a strong Seq2Seq model, which achieves state-of-the-art performance on both the Math23K and Ape210K datasets. However, these models highly rely on a method that extracting numbers from the question, and then mapping numbers to the slots of the generated templates. Such a two-stage process will introduce some systematic errors to the model. 

Therefore, we consider exploring the pipeline of generating expression directly instead of utilizing the template as an intermediate process, in which the model may gain more information from the question description and benefit from the end-to-end training strategy. 

\begin{figure*}[th]
  \centering
  \includegraphics[width=0.98\textwidth]{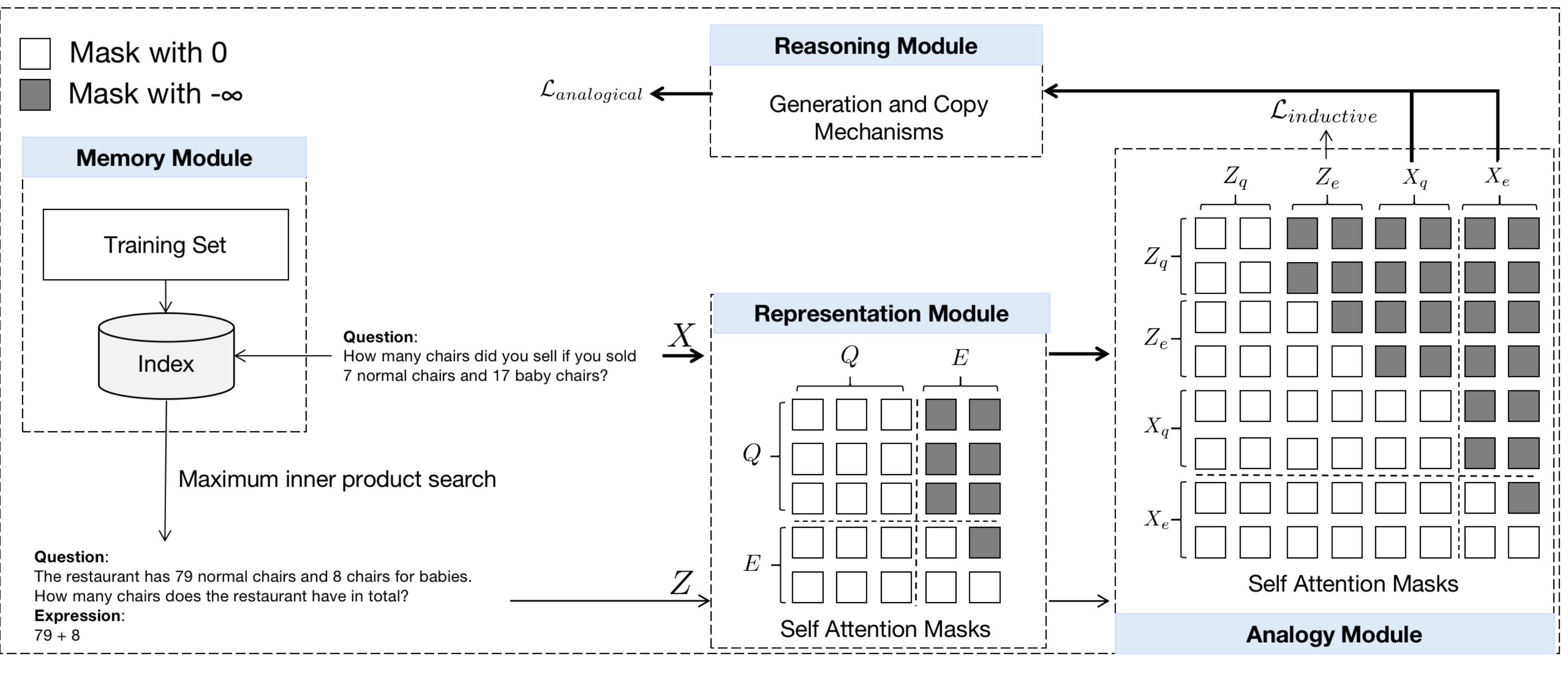}
  \vspace{-0.2cm}
  \caption{The illustration of our proposed REAL framework, which is composed of modules of memory, representation, analogy and reasoning. For an unsolved problem $X$, we first use maximum inner product search to find a similar question $Z$ from the memory module. And then the solution is generated with the copy mechanism in an analogical manner.}
  \label{fig:framework}
\end{figure*}

\subsection{Memory-augmented Generative Methods}
In the text generation task, there are mainly two types of models, one is based on retrieval \cite{zhou2016multi, zhou2018multi, zhang2018modeling, chen-etal-2019-bidirectional, wang2019multi}, and the other is based on generation \cite{qian2017assigning, zhou2018mojitalk, dong2019unified, han2019deep}. The retrieval algorithm can solve a particular task by constructing a knowledge base, which has high scalability. However, the retrieval-based approach cannot generate arbitrary results, which restricts the generation space of the model. In addition, the generative framework is able to store the knowledge in the model with the form of parameters, which has a certain generalization ability. However, in the knowledge-intensive task, it is difficult to remember all the knowledge in the parameters of generative model. To this end, many researchers attempted to combine retrieval and generation methods for text generation task \citep{zhang2017flexible, zhu2019retrieval, lewis2020retrieval, koncel2019text, chen2019towards, zhou2020improving}. In particular, \citet{zhang2017flexible} proposed a memory-augmented neural model for Chinese poetry generation, which investigates the contribution of memory. \citet{zhu2019retrieval} have demonstrated a retrieval-enhanced response generation approach for a dialogue system, which makes use of informative content in retrieved results to generate new responses. \citet{lewis2020retrieval} proposed a retrieval-augmented generation method where the parametric memory is a pre-trained Seq2Seq model and the non-parametric memory is a pre-trained neural retriever.

Our work is inspired by the success of incorporating memory into the generative model, showing memory-augmented model is capable of achieving strong performance in MWPs. Moreover, with the help of the memory module, our proposed model is able to solve the MWPs by analogy, which opens up a new research direction on MWP task.

\section{Method}
The framework of REAL is presented in Figure \ref{fig:framework}. In general, our proposed framework is composed of four key components: 1) Memory Module is constructed with a pre-trained model and is able to return top-K similar questions given a math word problem. 2) Representation Module is used to represent each token of the problem and the question in an inductive manner. 3) Analogy Module is utilized to aggregate the information of the problem and the retrieved question for better generating the correct expression. 4) Reasoning Module is combined with a copy mechanism that acts as a decoder to generate each token of the expression based on the input sequence.

\subsection{Problem Formulation}
We denote a problem as $X = \{X_q, X_e\}$, where the subscript $q$ and $e$ indicate the question description and mathematical expression respectively. $X_q$ is a sequence of word tokens $X_q = \{x_q^1, x_q^2, \cdots, x_q^L\}$, where $L$ is the length of the question description. We let its $K$ retrieved similar questions $\mathcal{Z} = \{Z^1, Z^2, \cdots, Z^K\}$ where $Z^i = \{Z_q^i, Z_e^i\}$. For each unsolved problem, the goal is to predict the token of $X_e$ at each time step $t$, namely $y_t \in \mathcal{V} \cup X_q$, where $\mathcal{V}$ is a generated vocabulary.

\subsection{Memory Module}
Aiming at solving a math word problem based on its similar retrieved questions, we employ a memory module to acquire external knowledge for enhancing the learning ability of the unsolved problem. The memory module is a non-parameter retriever, which is defined as the following formulation:
\begin{equation}
	p(Z|X) \approx p(Z_q|X_q) = \frac{e^{f(X_q)^{T}f(Z_q)}}{\sum_{Z_q}e^{f(X_q)^{T}f(Z_q)}},
\end{equation}
where $f(\cdot)$ is a Word2Vec \cite{mikolov2013efficient} model followed by a mean pooling technique that can represent a question description as a dense vector. In order to retrieve the similar question $Z_q$ given an unsolved problem $X_q$, we first normalize each vector and perform the MIPS (maximum inner product search) algorithm, which is implemented similar to the FAISS library \cite{JDH17}. Note that we utilize $p(Z_q|X_q)$ to approximate $p(Z|X)$ because only the problem description is provided in the testing stage.
% TODO: what's the effect of the casual mask?

\subsection{Representation Module}
The representation module is leveraged to summarize the representation of the problem and each retrieved question, which is called item memory. The module is constructed by the Transformer \cite{vaswani2017attention} block with a casual mask that similar to the settings of UniLM \cite{dong2019unified}, which can learn a bidirectional encoder and a unidirectional decoder simultaneously. Specifically, we perform a causal masking mechanism to allow each position in the expression to attend to previous positions, which preserve the auto-regressive property during decoding. In addition, we realize the representation of each token by summing the token, segment and position embeddings, which is similar to the approach of BERT model \cite{devlin2019bert}.
Next, follows the settings of UniLM \cite{dong2019unified}, to avoid the information-leakage problem during training, we use causal masks to ensure that the representation of each token in expression is only related to the previous states, as shown in Figure \ref{fig:framework}.

Therefore, in the training stage, given a problem $\{X_q, X_e\}$ with its corresponding retrieved questions $\{Z_q, Z_e\}$, the representation module is employed to acquire the item memories $\mathbf{X}_q$, $\mathbf{X}_e$, $\mathbf{Z}_q$ and $\mathbf{Z}_e$ with the same dimension of 768 respectively, which efficiently learns the representations of the problem and each retrieved question in an inductive manner.

\begin{figure}[t]
  \centering
  \includegraphics[width=0.475\textwidth]{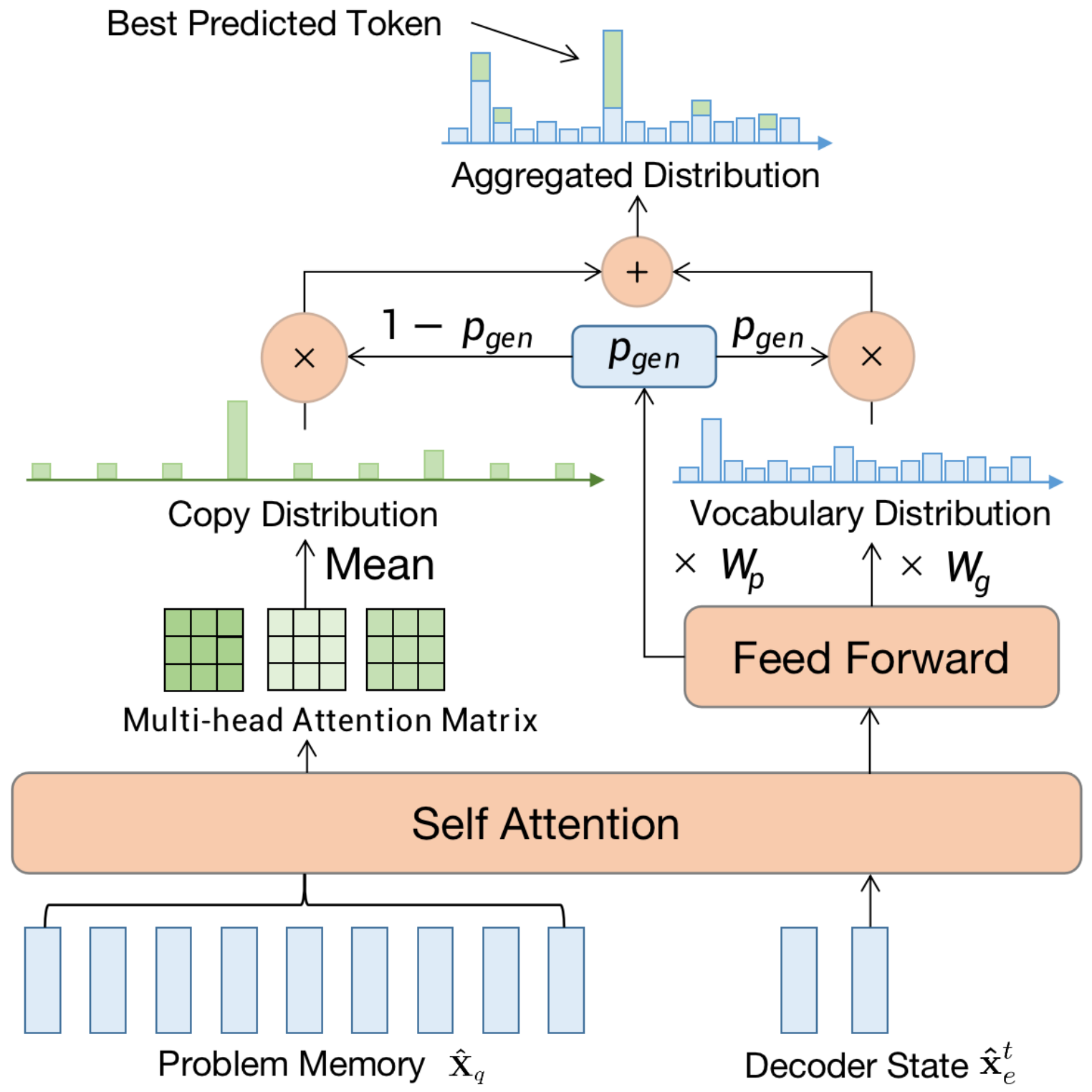}
  \caption{The overview of the reasoning module with a copy mechanism.}
  \label{fig:copymechanism}
\end{figure}

\subsection{Analogy Module}
In order to achieve the way of analogical learning, the model needs to aggregate contextual information from the item memories of the unsolved problem and the retrieved questions. Therefore, we first concatenate item memories to form input features $\{\mathbf{Z}_q, \mathbf{Z}_e, \mathbf{X}_q, \mathbf{X}_e\}$ and preprocess the input features using the mechanisms of position encoding and segment encoding. Thereafter, based on the length of the input sequences $Z_q$, $Z_e$, $X_q$ and $X_e$, a casual mask can be constructed similar to the approach of representation module, as shown in Figure \ref{fig:framework}. The purpose of the casual mask is to enhance the analogical learning capability by focusing the attention of unsolved problem on the retrieved questions. In addition, the expression part in a casual mask is designed to only attend to the previous token, which avoids the information-leakage problem. Lastly, we utilize a Transformer network that similar to the representation module for learning relational memories by analogy. Thereinto, the output states of the analogy module are denoted as relational memories $\mathbf{\hat{Z}}_q$, $\mathbf{\hat{Z}}_e$, $\mathbf{\hat{X}}_q$ and $\mathbf{\hat{X}}_e$ respectively. Note that $\mathbf{\hat{Z}}_q$ and $\mathbf{\hat{Z}}_e$ are the outputs of the last layer, and the $\mathbf{\hat{X}}_q$ and $\mathbf{\hat{X}}_e$ are the outputs of penultimate layers.

In order to extract the knowledge from the retrieved question $\{Z_q, Z_e\}$, we further employ a classifier $C \in \mathbb{R}^{768 \times |V|}$ to solve the question description of $Z_q$ and propose an auxiliary loss $\mathcal{L}_{inductive}$ to navigate the learning direction of the analogy module, which is formulated as follows:
\begin{equation}
	\mathcal{L}_{inductive} = -\sum_{t}^{N}log p_{\theta_{a}}(z_e^t|Z_q, z_e^{1:t-1}),
	\label{equ: inductive loss}
\end{equation}
where $z_e^t$ indicates the $t^{th}$ token of $Z_e$, and $\theta_{a}$ represents the parameters of analogy module.

\subsection{Reasoning Module}
Taking the structure of the word math problem into account, we know that the operands of an expression are likely to come from the problem description $X_q$. Therefore, we design a reasoning module with a copy mechanism \citep{see2017get}, which is build based on the last layer of analogy module. As shown in Figure \ref{fig:copymechanism}, given a decoder state $\mathbf{\hat{x}}_e^t$, the vocabulary distribution $p_g(y_t|X_q)$ and the copy distribution $p_c(y_t|X_q)$ are formulated as follows:
\begin{gather}
	\begin{align}
		p_g(y_t|X_q) &= \frac{e^{\phi_g(y_t)}}{\sum_{y \in V}e^{\phi_g(y)}}, \\
		p_c(y_t|X_q) &= \frac{1}{h}\sum_{j \le h} \sum_{i:x_i=y_t}\frac{\phi^{j}_x(x_i)}{\sum_{x_k \in X_q}\phi^{j}_x(x_k)},
	\end{align}
\end{gather}
where the generated probability $p_g(y_t| X_q)$ is implemented as a fully-connected layer $\phi_{g}$ followed by the analogy module with weights $W_g$. And $\phi^{j}_x(x_i)$ indicates the $j^{th}$ head attention value \citep{vaswani2017attention} of token $x_i$, $h$ is the total number of the attention head.

To combine the vocabulary distribution $p_g(y_t|X_q)$ and the copy distribution $p_c(y_t|X_q)$, we use a learnable value $p_{gen}$ to calculate the aggregated distribution $p(y_t|X_q)$ as follows:
\begin{equation}
	 p_{gen}p_g(y_t|X_q)+(1-p_{gen})p_c(y_t|X_q),
\end{equation}
where probability $p_{gen}$ is computed by a fully-connected layer followed by the analogy module with weights $W_p$. Therefore, the reasoning module can decide whether to copy the number in the problem description according to the context.

\subsection{Learning Details}
Suppose the length of expression of a problem is $N$, the goal of our model is to generate a token probability distribution $p_{\theta}(y_t|X_q, Z, y_{1:t-1})$ based on the problem and its retrieved question, where $t \le N$ and $\theta$ is the parameters of the model. Next, we marginalize the token distribution to generate the $t^{th}$ output distribution $p_{\theta}(y_t|X_q, y_{1:t-1})$ based on the top-K retrieved questions $Z$. Finally, generating each token $y_t$ sequentially is able to form a complete expression $X_e$ of problem $X_q$. Formally, the framework $p_{\theta}(y|X_q)$ can be defined as follows:
\begin{equation}
	 \prod_{t}^{N} \mathop{\mathbb{E}}_{Z \in top-K(p(Z|X))} p_{\theta}(y_t|X_q, Z, y_{1:t-1}),
	\label{eqn: framework}
\end{equation}
where $top{-}K(p(Z|X))$ is a probability model that instantiated as a memory module to retrieve K similar questions. 
The loss function can be defined as the negative marginal log-likelihood as follows:
\begin{equation}
	\mathcal{L}_{analogical} = -log(p_{\theta}(y|X_q)),
	\label{equ: analogical loss}
\end{equation}
where $p_{\theta}(y|X_q)$ is a probability model of REAL illustrated in Eqn. (\ref{eqn: framework}). In order to facilitate the inductive learning of model, we further employ an auxiliary loss illustrated in Eqn. (\ref{equ: inductive loss}). Therefore, the total loss function is defined as a weighted sum of analogical loss and inductive loss. Formally, our training goal is formulated as follows:
\begin{equation}
	\mathcal{L} = \mathcal{L}_{analogical} + \lambda \mathcal{L}_{inductive},
	\label{equ: combination loss}
\end{equation}
which $\lambda$ is a hyperparameter for balancing the weights between $\mathcal{L}_{analogical}$ and $\mathcal{L}_{inductive}$. We simply set $\lambda$ equal to 1 and found it works well in all experiments. 

\section{Experiments}
In this section, we conduct extensive experiments on two well-known datasets to answer the following five research questions:
\begin{itemize}
	\item [\textbf{RQ1}] How does our proposed REAL framework perform as compared to other state-of-the-art competitors?
	\item [\textbf{RQ2}] Are memory and copy mechanisms equally important? How does REAL model perform if one mechanism is removed?
	\item [\textbf{RQ3}] How does REAL perform with respect to various number of retrieved questions?
	\item [\textbf{RQ4}] How does REAL perform when solving problems of varying expression lengths (difficulties)?
	\item [\textbf{RQ5}] Can we visualize the solving process for MWP task?
\end{itemize}

\subsection{Experimental Settings}
\subsubsection{Datasets}
We evaluate our framework on two datasets, Math23K\footnote{\url{https://github.com/SumbeeLei/Math_EN/tree/master/data}} \cite{wang2017deep} and Ape210K\footnote{\url{https://github.com/Chenny0808/ape210k}} \cite{zhao2020ape210k}. The Math23K dataset labeled with equations and answers contains 22,162 questions in the training set and 1,000 questions in the testing set.
Since most of the state-of-the-art results were experimented via 5-fold cross-validation and a published testing dataset, we evaluate REAL on both settings. 
Ape210K is a relatively large-scale dataset containing 210,488 math word problems, which are split into training, validation and testing subsets. Both validation and testing subsets have 5,000 samples and we leave the rest of 200,488 as the training samples. 

\subsubsection{Baselines}
To justify the effectiveness of our method, we compare it to state-of-the-art baselines:
\begin{itemize}
	\item \textbf{DNS \citep{wang2017deep}.} This is a vanilla Seq2Seq model that jointly utilizes a number mapping technique and an equation template technique to generate the expression of problems.
	\item \textbf{Math-EN \citep{wang2018translating}.} A preprocessed technique that called equation normalization is proposed to significantly reduce the template space. 
	\item \textbf{T-RNN \citep{wang2019template}.} This method applies a tree-structure Seq2Seq model to predict suffix expression, with inferred numbers as leaf nodes and unknown operators as inner nodes.
	\item \textbf{StackDecoder \citep{chiang2019semantically}.} This method proposes a stack-based decoding process to model semantic meanings of operands and operations of MWPs.
	\item \textbf{GTS \citep{xie2019goal}.} This is a goal-driven tree-structured model to decode the expression in both top-down and bottom-up manners.
	\item \textbf{TSN-MD \citep{2020Teacher}.} This method proposes a teacher-student networks with multiple decoders to improve the diversity of generated expressions.
	\item \textbf{Graph2Tree \citep{zhang2020graph}.} This method designs a graph network to enrich quantity representations and decodes the expression using a tree-based decoder like GTS.
	\item \textbf{Ape \citep{zhao2020ape210k}.} This paper proposes a feature-enriched and copy-augmented Seq2Seq model, which achieves competitive performance on both Math23K and Ape210K datasets. 
\end{itemize}

\begin{table}[t]
\centering
	\setlength{\tabcolsep}{1.1mm}{
	\begin{tabular}{c|c|c|c}
	\hline
	Model & Math23K & Math23K* & Ape210K \\ 
	\hline
	DNS 	& - 	& 58.1	& 	-	\\
	Math-EN & 66.7 	& -		&	-	\\
	T-RNN 	& 66.9 	& -		& -	\\
	StackDecoder &-	& 65.8	& 52.28	 	\\
	GTS  	& 75.6 & 74.3 		& 56.56		\\
	TSN-MD 	& 77.4 & 75.1  	& -		\\
	Graph2Tree & 77.4 & 75.5&	-		\\
	Ape &	-	&	77.5	& 70.20 	\\ \hline
	REAL & \textbf{82.3} & \textbf{80.8}	& \textbf{77.18}  \\
	\hline
	\end{tabular}}
	\caption{The overall comparison of REAL and various methods on Math23K and Ape210K datasets. Note that Math23K denotes results on public testing set and Math23K* denotes 5-fold cross-validation. Note that the previous results evaluated on Ape210K dataset are published by \citet{zhao2020ape210k}. (Section \ref{section: RQ1})}
	\label{table: oc}
\end{table}

\subsubsection{Implementation Details}
Our model is implemented based on the PyTorch\footnote{\url{http://www.pytorch.org}} framework on a server equipped with 2 NVIDIA 1080Ti GPU. In the REAL model, the representation module and analogy module are both constructed by 6 layers Transformer block \cite{vaswani2017attention}. To initialize the hidden layers in Transformer, we set their parameters with a pre-trained BERT \cite{devlin2019bert}. The equation normalization technique \citep{wang2018translating} is applied in the training stage, which follows the previous works for fair comparison. Our model is trained for 80 epochs where the mini-batch size is set to 12. In each mini-batch, problems with their corresponding retrieved questions are randomly sampled from the training set. For optimizer, we use ADAM optimization algorithm \cite{kingma2014adam} with the learning rate of 5e-4, $\beta_1=0.9$ and $\beta_2=0.99$. In addition, the learning rate is halved per 5 epochs when the total epoch is greater than 40 and we also set beam size to 5 in beam search during decoding. Lastly, we treat the predicted expression as correct if its calculated value equals to the answer, and we use the answer accuracy as the evaluation metric which follows previous works \citep{wang2018translating, zhao2020ape210k}.

\begin{table}[t]
\centering
	\setlength{\tabcolsep}{4.0mm}{
	\begin{tabular}{c|c|c}
	\hline
	\textbf{Model} & Math23K* & Ape210K \\ 
	\hline
	REAL & 80.8 & 77.18 \\
	\hline
	w/o EN & - 0.6 & - 0.16 \\
	\hline
	w/o Copy & - 0.4 & - 0.56 \\
	\hline
	w/o Memory & - 0.9 & - 0.62 \\
	\hline
	w/o All & - 1.6 & - 0.70 \\
	\hline
	\end{tabular}}
	\caption{Performance comparisons of various components on Math23K* and Ape210K datasets. (Section \ref{section: RQ2})}
	\label{table: as}
\end{table}

\subsection{Overall Performance Comparison (RQ1)} \label{section: RQ1}
To demonstrate the effectiveness of our proposed REAL solution, we compare it to several state-of-the-art approaches: 1) DNS; 2) Math-EN; 3) T-RNN; 4) StackDecoder; 5) GTS; 6) TSN-MD; 7) Graph2Tree; and 8) Ape.

Table \ref{table: oc} shows the comparison results on Math23K and Ape210K datasets among different methods, we have the following observations: 1) Our proposed REAL method shows the best performance on all benchmark datasets as compared to other methods. To verify the statistical significance of our improvement, we further conduct one-sample t-test on Math23K* experiments compared to the accuracy of Ape model and acquire a p-value about 4e-4, which unveils the superiority of our algorithm. 2) Jointly observing the experimental results on Math23K and Ape210K, we can see that our proposed model has better improvement on Ape210K dataset as compared to the improvement on Math23K. This is probably because our model is more effective on the large-scale dataset. 3) We do not perform any handcraft preprocessing steps to reduce the difficulty of model training, such as number mapping \cite{wang2017deep, zhao2020ape210k} and relation extraction \cite{zhang2020graph}, and still achieves great performance, which manifests the effectiveness of our proposed framework.

\subsection{Ablation Study (RQ2)} \label{section: RQ2}
To evaluate the effectiveness of our proposed analogical learning method, especially the design of equation normalization technique, memory component and copy mechanism, we conduct ablation study on these components. In particular, we employ EN to denote equation normalization technique, Copy to denote the copy mechanism and w/o Memory to denote the model trained by inductive loss without using the memory module.

The performance of the three-component ablation study is shown in Table \ref{table: as}. We have the following observations: 
1) By comparing the results of REAL and w/o EN, the performance of model is benefited from the equation normalization technique, which reveals its effectiveness in MWP task.
2) Jointly observing the performance of w/o Copy and w/o Memory models, we can infer that the Memory component is more important than the Copy component. This is mainly because the memory component is the key to perform analogical learning, which can learn the intrinsic relationships among unsolved problem and similar questions. Meanwhile, the copy component is reasonable due to it takes the structure of MWP task into account, which also results in better performance.  
3) By comparing the results of w/o All with the other experiments, we find the accuracy drops significantly, proving that the three components have positive impacts on the model's performance consistently.

\begin{figure}[t]
  \centering
  \includegraphics[width=0.475\textwidth]{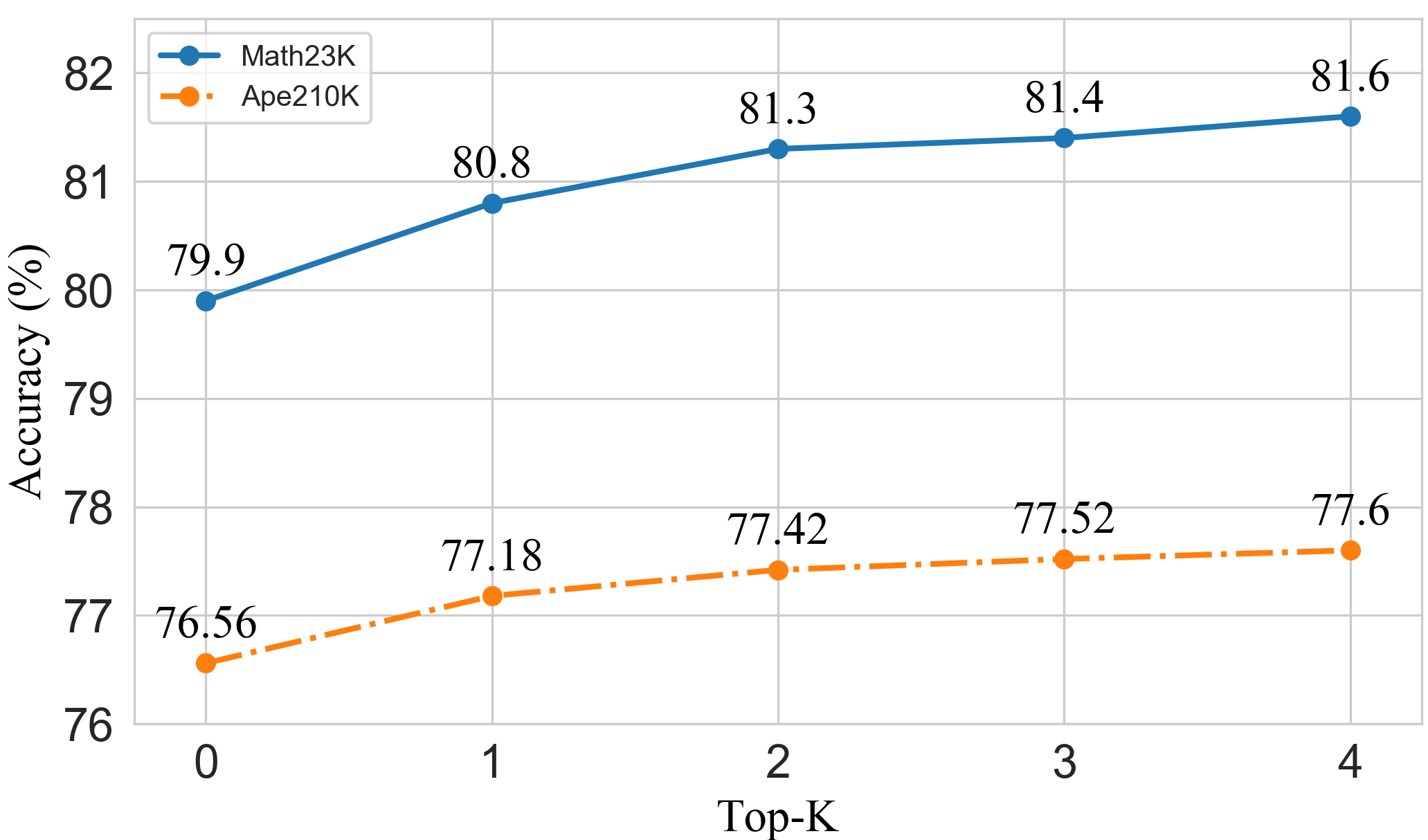}
  \caption{The performance of REAL w.r.t. various number of retrieved questions. (Section \ref{section: RQ3})}
  \label{fig: top-K}
\end{figure}

\subsection{Impact of Retrieved Questions (RQ3)} \label{section: RQ3}
Although REAL is trained with only a retrieved question, we still have the flexibility to adjust the number of retrieved questions at the testing stage, which can affect the model's performance. In order to show that REAL is able to solve MWPs by analogy, we test the model according to various number of retrieved questions on Math23K* and Ape210K datasets. 

As shown in Figure \ref{fig: top-K}, we have the following observations: 
1) With the increased number of the retrieved questions, the model's performance is monotonically improved. This clearly shows that REAL model is able to master the knowledge in an analogical way, which manifests the rationality of our proposed framework. 
2) It is obviously observed that when K increases from 0 to 1, the model's performance achieves significant improvement. This is mainly because the training method of the model is changed from an inductive way to an analogical way, showing the effectiveness of the memory components. 
3) The performance on both datasets are relatively stable and reach their maximum values when $K=4$. It indicates that with the increase of K, the marginal benefits of improving the model's performance will gradually diminish. We consider the noise introduced by the retrieved questions may affect the performance. Because the more questions retrieved by the memory module, the lower the similarity of the corresponding questions. 4) The experimental results demonstrate REAL's flexibility in balancing the performance and efficiency, which is an advantage of performing our memory-augmented framework in practical application.

\begin{figure}[t]
  \centering
  \includegraphics[width=0.49\textwidth]{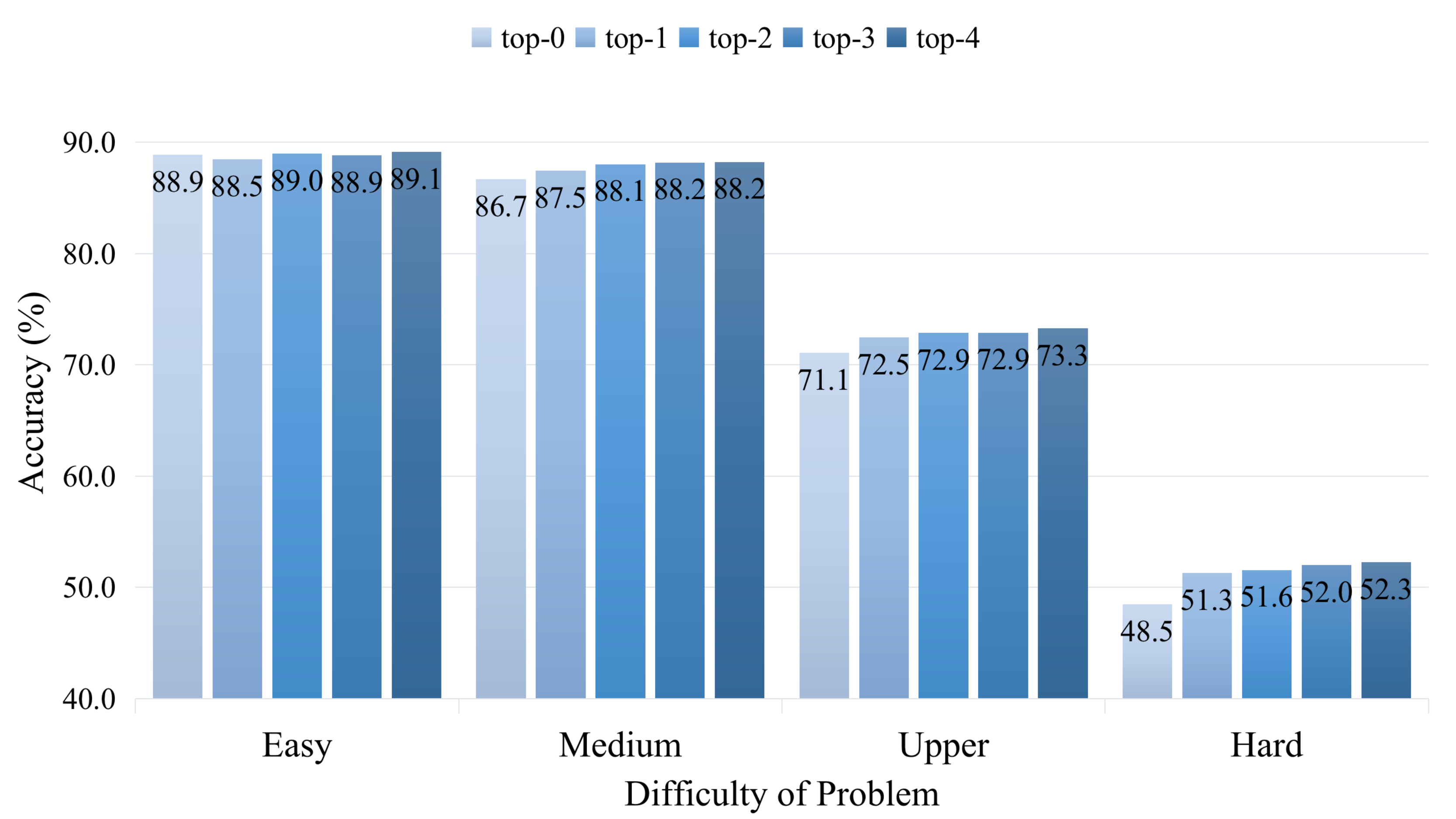}
  \vspace{-0.5cm}
  \caption{The top-K performance of REAL w.r.t. varying difficulties of unsolved problems. (Section \ref{section: RQ4})}
  \label{fig: diff}
\end{figure}

\subsection{Impact of Length (Difficulty) (RQ4)} \label{section: RQ4}
To further evaluate REAL's analogy ability based on MWPs with different difficulty, we split every fold of Math23K dataset into 4 subsets according to the length of expression. Specifically, we deem that the longer the length of expression, the more difficult the corresponding problem is, and vice versa. Therefore, we sort the problems of the testing set according to the length of the expression in ascending order, and split it into 4 subsets, which are categorized as different difficulty levels of easy, medium, upper and hard. According to this, we conduct 20 experiments to consider how the retrieved number of questions will affect the performance under different difficulty problems. Note that each experimental result is obtained by averaging the results of the 5-fold subsets. 

As shown in Figure \ref{fig: diff}, we have the following observations: 
1) With the increase of difficulty, the performance of REAL gradually decreases, which is reasonable because the longer the length of expression, the more difficult for the model to predict. 
2) In the ``Medium'', ``Upper'' and ``Hard'' experiments, the analogical results of $K \geq 1$ are noticeably superior to the inductive results of $K = 0$, which manifests the rationality of analogical learning method. Furthermore, as the number of retrieved memories increases, the model's performance is consistently improved. It demonstrates the effectiveness of our proposed analogical learning method. 
3) In contrast, the experimental results are unstable when the difficulty of the problem is ``Easy''. We consider the reasons behind are: a) The solutions of simple problems with shorter expressions are easy to master by the model, so the model is far more likely to rely on the inductive method and can not benifit from more analogies. b) When solving relatively easy problems, the performance of the inductive-preferred model may be harmed due to the noise introduced by the increased retrieved questions. 
This indicates the quality of retrieved questions should be carefully considered and we leave it for future research.

\begin{table*}[t]
	\centering
	\begin{tabular}{|l|p{0.1\textwidth}|p{0.73\textwidth}|}
	\hline
	\multirow{5}{*}{Case 1} & Problem                           & To build a swimming pool with a length of 18 meters, a width of 10 meters, and a depth of 2 meters. We need to surface the walls and bottom of the swimming pool with cement, how many square meters of cement should be applied?                                      \\ \cline{2-3} 
						   & Inductive Prediction               & ( 18 $\times$ 10 $+$ 10 $\times$ 2 $+$ 2 $\times$ 2 ) $-$ 18 $\times$ 10  {\color{red}\xmark}                                                                                                                                                      \\ \cline{2-3}
						   & \multirow{2}{*}{\shortstack{Retrieved\\Questions}} & 1) \emph{Question:} A rectangular swimming pool is 60 meters long, 40 meters wide, and 2 meters deep. Now we need to put cement on the walls and bottom. What is the area of the cement? \newline \emph{Equation:} ( 60 $\times$ 40 $+$ 40 $\times$ 2 $+$ 60 $\times$ 2 ) $\times$ 2 $-$ 60 $\times$ 40
						   \\ \cline{3-3} 
						   &                                    & 2) \emph{Question:} A rectangular water pool, 20 meters long, 10 meters wide, and 2 meters high. We need to surface the walls and bottom of the pool with cement. How many square meters of cement do we need to apply? \newline \emph{Equation:} 20 $\times$ 10 $+$ 20 $\times$ 2 $\times$ 2 $+$ 10 $\times$ 2 $\times$ 2 \\ \cline{2-3} 
						   & Analogical Prediction                 & ( 18 $\times$ 10 $+$ 10 $\times$ 2 $+$ 18 $\times$ 2) $\times$ 2 $-$ 18 $\times$ 10 {\color{green}\checkmark}                                                                                                                                                     \\ \hline
	\multirow{5}{*}{Case 2} & Problem                           & A class held a math competition with a total of 20 questions. It is stipulated that 5 points will be given for one correct answer, and 2 points will be deducted for one wrong answer. Xiao Ming got 86 points. How many questions did he answer correctly?                                                                                                             \\ \cline{2-3} 
						   & Inductive Prediction               &  ( 20 $\times$ 5 $-$ 86 ) $\div$ ( 5 $+$ 2 ) {\color{red}\xmark}                                                                                                                                                             \\ \cline{2-3} 
						   & \multirow{2}{*}{\shortstack{Retrieved\\Questions}} & 1) \emph{Question:} There are 20 questions in total. 7 points will be given for one correct answer, and 4 points will be deducted for one wrong answer. Wang Lei scored 74 points. How many questions did he answer correctly? \newline \emph{Equation:} 20 $-$ ( 20 $\times$ 7 $-$ 74 ) $\div$ ( 7 $+$ 4 )                                                                                                    \\ \cline{3-3} 
						   &                                    & 2) \emph{Question:} In the knowledge competition, there are 10 judgment questions. The scoring rules are: 2 points for each correct answer, and 1 point will be deducted for wrong answer. Xiao Ming only got 14 points. How many questions did he answer correctly? \newline \emph{Equation:}  ( 14 $+$ 10 $\times$ 1 ) $\div$ ( 2 $+$ 1 )                                                                                                            \\ \cline{2-3} 
						   & Analogical Prediction                 &  20 $-$ ( 20 $\times$ 5 $-$ 86 ) $\div$ ( 5 $+$ 2 )  {\color{green}\checkmark}                                                                                                                                                         \\ \hline
	\end{tabular}
	\caption{Two cases of REAL solving MWPs using inductive mode and analogical mode. (Section \ref{section: cs})}
	\label{table: cs}
	\end{table*}
	
	\subsection{Case Study (RQ5)} \label{section: cs}
	To better understand how the analogical learning method work in MWP task, we exploited some  macro-level case studies. Specifically, we first trained a REAL model with Top-2 settings in the Math23K dataset, and selected two hard problems from the testing set that can not be solved in inductive mode but solve correctly by the analogical one.
	
	As shown in Table \ref{table: cs}, the case 1 describes a problem about surfacing a swimming pool by cement. The prediction is wrong when the model try to solve the problem in an inductive manner. It seems that the model is lack of common sense about the formula of cube area and misunderstands the concept of depth. To this end, we attempt to solve the problem using analogical method, which results in a correct solution. From the descriptions of problem and the retrieved questions, we can see that the REAL model is able to discover the common structure among the problem and the retrieved questions, and solve the problem through the expression template of the retrieved questions in an analogical manner. Case 2 describes a counting problem that the quantitative relationship is very complicated, in which an ingenious and complex reasoning process is required for solving the problem correctly. As shown in Table \ref{table: cs}, it is as expected that our model fail to solve this complex problem in an inductive manner, because the existing deep learning models are still difficult to have human-like reasoning ability. In constrast, the analogical one can generate a correct solution by refering to the similar questions, which demonstrates that our proposed framework is able to learn by analogy. 
	
	The above two cases qualitatively show that the memory-augmented component is an effective structure in REAL framework, which introduces an novel analogical approach for MWP task and opens a new possibility for future work.

\section{Conclusion And Future Work}
In this work, we propose a memory-augmented solver called REAL for MWPs. Under the REAL framework, there are four key components: 1) Memory module; 2) Representation module; 3) Analogy module; 4) Reasoning module, which are proposed to perform analogical learning schema based on the retrieved similar questions. In addition, to enhance the generation performance, a copy mechanism is designed to properly aggregate the information of operands from the problem description. The experimental results show that REAL achieves state-of-the-art performance for MWP task. Extensive micro-scope studies demonstrate the ability of REAL in learning by analogy.

In the future, we plan to extend our work in the following two directions. First, the model's performance can be further improved if the memory module of REAL model is jointly trained with the whole framework. Second, we will consider designing a more meaningful analogy module that can take the structure of question and expression into account, thus providing more information for the reasoning module to generate the problem solution.

\bibliography{paper}
\bibliographystyle{acl_natbib}

%\appendix
%
%\section{Example Appendix}
%\label{sec:appendix}
%
%This is an appendix.

\end{document}